\documentclass{article}
\usepackage{authblk}
\usepackage[margin=1in]{geometry}
\usepackage{graphicx}
\usepackage{amsmath,amssymb}
\usepackage[square,numbers]{natbib}
\usepackage{hyperref}

\title{Information-Seeking Decision Strategies Mitigate Risk in Dynamic, Uncertain Environments}

\author[1]{Nicholas W. Barendregt}
\affil[1]{BioFrontiers Institute, University of Colorado Boulder}
\author[2]{Joshua I. Gold}
\affil[2]{Department of Neuroscience, University of Pennsylvania}
\author[3]{Kre\v simir Josi\' c}
\affil[3]{Departments of Mathematics, Biology and Biochemistry
University of Houston}
\author[4]{Zachary P. Kilpatrick}
\affil[4]{Department of Applied Mathematics, University of Colorado Boulder}
\date{}

\begin{document}
\maketitle
\begin{abstract}
    To survive in dynamic and uncertain environments, individuals must develop effective decision strategies that balance information gathering and decision commitment.
    Models of such strategies often prioritize either optimizing tangible payoffs, like reward rate, or gathering information to support a diversity of (possibly unknown) objectives.
    However, our understanding of the relative merits of these two approaches remains incomplete, in part because direct comparisons have been limited to idealized, static environments that lack the dynamic complexity of the real world. 
    Here we compared the performance of normative reward- and information-seeking strategies in a dynamic foraging task. 
    Both strategies show similar transitions between exploratory and exploitative behaviors as environmental uncertainty changes. 
    However, we find subtle disparities in the actions they take, resulting in meaningful performance differences: whereas reward-seeking strategies generate slightly more reward on average, information-seeking strategies provide more consistent and predictable outcomes. 
    Our findings support the adaptive value of information-seeking behaviors that can mitigate risk with minimal reward loss.
\end{abstract}
\section{Introduction}
    Effective behavioral strategies are adaptive~\cite{gilmour2018plasticity,ropert2009impact}. 
    This adaptivity is often cast in terms of normative frameworks, which use objective functions that represent preferences in both actions and outcomes~\cite{barendregt2022, bogacz2006, drugowitsch2012, nguyen2019, tajima2019}. 
    However, the objectives that drive organisms' natural behaviors can be complex and remain poorly understood. 
    One challenge is that strategies with different objectives can produce similar behaviors. 
    For example, it can be difficult to distinguish a strategy that aims to maximize a particular outcome, like reward rate, from one that aims to maximize information that is useful for achieving that outcome. 
    This challenge is particularly acute in naturalistic settings, where there is often a complex relationship between particular outcomes and information that is relevant to those outcomes.
    
    This interplay between rewards and information is evident in strategies for resource foraging. 
    Foraging typically requires sequential decisions that alternate between choosing to gather evidence (e.g.,~about local resource availability) and using this information to take actions (e.g.,~keep exploiting local resources or explore more broadly for other resources)~\cite{hayden2014, hills2006, mobbs2018}. 
    Many theoretical accounts of foraging have focused on the optimization of a reward-based objective.  
    A notable example is the pervasive marginal value theorem (MVT) which is based on optimizing the average rate of reward (e.g., food/resource) intake~\cite{charnov1976,pyke1984optimal}. 
    These reward-optimal approaches have provided key insights into foraging behaviors~\cite{green1984stopping, valone2006animals, kilpatrick2021, mcnamara1982,mobbs2018}. 
    However, debates continue about whether animals are truly reward-rational~\cite{olschewski2025} or even capable of implementing truly normative reward-maximizing strategies~\cite{cisek2009, ho2022}.
    
    Alternatively, animals could approximate reward optimization by appealing to other intermediate objectives that serve this outcome in flexible ways~\cite{bartumeus2016foraging}. 
    One class of such strategies is information maximization, which drives foragers to become as knowledgeable as they can about the environment. 
    These strategies do not optimize explicitly for a tangible outcome, such as immediate reward intake, but can indirectly produce reliably favorable outcomes, accommodate environmental changes and uncertainty, and be less error prone than fine-tuned normative strategies. 
    One example of an information-maximizing strategy is infotaxis, in which individuals move to locations they expect to provide the largest reduction in their uncertainty about the environment~\cite{vergassola2007}. 
    This strategy is particularly effective for source-tracking tasks~\cite{barbieri2011, karpas2017}, producing behavioral trajectories that qualitatively replicate experimental findings~\cite{calhoun2014, voges2014}.
    However, direct comparisons between information-seeking strategies like infotaxis and reward-seeking strategies like those based on the MVT are rare~\cite{loisy2022} and are typically restricted to search in static environments. 
    How these model classes compare in more naturalistic (e.g., dynamic, sequential, uncertain) settings remains unknown.
	
    We compared how sequential versions of normative reward-maximizing (``rewardmax'') and infotaxis (``infomax'') strategies perform in dynamic foraging environments. 
    Simulated agents performed a sequential, two-alternative foraging task that included  manipulations of both the probability that the correct alternative changes between trials and the probability that choosing the correct alternative yields a reward~\cite{nguyen2020}. 
    Our results show that both models predict qualitatively similar behaviors as the environmental uncertainty changes, recreating the explore-exploit trade-off observed in many classic foraging studies. 
    As expected, the rewardmax model maximizes average reward. 
    However, the infomax model obtains reward more reliably than the rewardmax model, thereby mitigating risk. 
    This reliability is consistent across both environmental and model perturbations. 
    Together, these results illustrate an important advantage of information maximization as a foraging strategy in naturalistic settings and point to relevant task designs where these two strategies can be further disentangled.
    
\section{Methods}
    \label{sec:methods}
    The rewardmax and infomax models each use the same (Bayesian) evidence-gathering process to update beliefs, obtained from the posterior probabilities about the environment. 
    The models differ in terms of the rules they use to terminate evidence gathering and commit to a choice. 
    Below we first describe a common evidence-gathering process evolving within and across trials, then describe the different commitment rules for each model that we obtain via dynamic programming.
    
    \subsection{Sequential Task Development and Evidence Integration}
        We consider a sequence of foraging decisions in an environment that, for each decision, takes on one of two possible hidden states, $s^i\in\{s_+,s_-\}$, where the superscript, $i$, denotes the $i$-th state in the sequence.
        Each state determines which of two options is more likely to provide a reward to a foraging agent. 
        For instance, the state could represent which one of two resource patches is more likely to yield edible plants~\cite{levin2000,mcmahon2017}.
        To infer the state of the environment, the agent makes independent and identically distributed observations, $\xi^i_j$, each of which follows a state-dependent distribution, $f_\pm\left(\xi^i_j\right)=f\left(\xi^i_j\vert s^i=s_\pm\right)$.
        A Bayesian agent deduces which state is more likely after $n$ observations by computing the log-likelihood ratio (LLR), $y_n^i$, which we will also refer to as the agent's ``belief.'' 
        The best possible estimate of the state from the sequence of noisy observations $\xi_{1:n}^i$~\cite{bogacz2006, gold2007, wald1945} is determined by computing:
        \begin{equation}y_n^i=\ln\frac{\Pr\left(s^i=s_+\vert\xi^{i}_{1:n}\right)}{\Pr\left(s^i=s_-\vert\xi^i_{1:n}\right)}=\ln\frac{f_+\left(\xi_n^i\right)}{f_-\left(\xi_n^i\right)}+y_{n-1}^i.
            \label{eq:within-trial LLR update}
        \end{equation}
        To obtain Eq.~\eqref{eq:within-trial LLR update}, we used the conditional independence of the observations $\xi_{1:n}^i$ to obtain an iterative update rule for the LLR. 
        This update rule involves a prior belief, $y_0^i$, carried forward from the previous trial when $i=2,3,...$, as described below. 
        When $i=1$, we assume $y_0^1 = 0$; both states are considered equally likely to be drawn as the initial state $s^1$.
		
\begin{figure}[t]
            \centering
            \includegraphics[width=\linewidth]{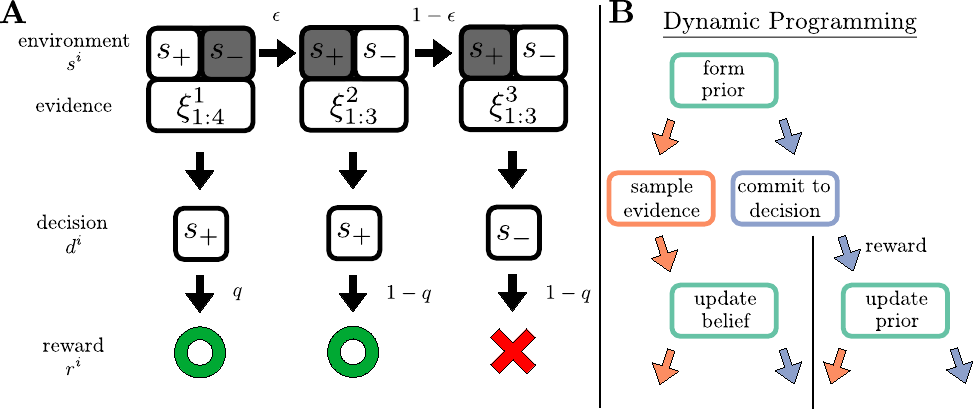}
            \caption{\textbf{Dynamic Foraging Task and Modeling Approach.}
                \textbf{A:} Schematic of an example realization of the dynamic foraging task.
                Between decisions, the environment is in a constant, hidden state $s^i$ that the agent infers from a sequence of observations, $\xi^i_{1:n}$.
                Once sufficiently confident, the agent makes decision $d^i$ and receives a probabilistic reward $r^i$ with reliability $q$.
                The environment then changes state with probability $\epsilon$, and the process begins again.
                \textbf{B:} Schematic of the dynamic programming approach for objective-maximizing behavior.
                At each time step the agent uses prior information to calculate the expected utility of sampling evidence and committing  to a decision.
                The agent then takes the action that maximizes expected utility (either reward or information) and updates their belief based on the information gained from the environment (in the case of sampling) or from probabilistic reward (in the case of commitment).}
            \label{fig:task_schematics}
        \end{figure}

        An agent's decision $d^i\in\{s_+,s_-\}$ about the hidden state $s^i$ triggers two events.
        First, the agent receives a probabilistic reward or punishment $r^i\in\{\circ,\times\}$ based on their choice, determined by a Bernoulli random variable conditioned on the correctness of the decision (correct, $d^i = s^i$, or incorrect $d^i \neq s^i$)~\cite{nguyen2020}:
        \begin{equation}
            \begin{gathered}
                \Pr\left(r^i=\circ\vert d^i=s^i\right)=\alpha,\\
                \Pr\left(r^i=\times\vert d^i\neq s^i\right)=\beta.
            \end{gathered}
            \label{eq:reward process}
        \end{equation}
        For convenience, we will take $\alpha=\beta=q$ throughout this work.
        Because $q$ is the probability the feedback matches the correctness of the agent's response, we can interpret this parameter as the reliability of patch resource yields.
        Second, once the agent makes a decision, the environment changes state with a fixed probability, $\epsilon$~\cite{nguyen2019}:
        \begin{equation}
            \Pr\left(s^{i+1}=s_\pm\vert s^i=s_\mp\right)=\epsilon.
            \label{eq:state transition process}
        \end{equation}
        This change probability captures the timescale of temporal fluctuations present in foraging environments, such as temperature changes that can make some resource patches more bountiful, or the appearance of predators that reduce the chances of safe and successful foraging.
        After feedback is delivered and the state updates, the $i+1$-st trial begins.  
        The observer may continue to make observations, starting with $\xi^{i+1}_1$, until they make decision $d^{i+1}$.
        An example realization of this process is schematized in Fig.~\ref{fig:task_schematics}\textbf{A}.
		
        Having defined the environmental feedback and transition processes in Eqs.~(\ref{eq:reward process}-\ref{eq:state transition process}), we can now give the expression for the updated prior, $y_0^{i+1},$ after the $i$-th decision.
        This prior depends on the belief $y_n^i$ that determined decision $d^i$, as well as the ensuing probabilistic reward, $r^i$.
        Using the same representation of the belief as the LLR used in Eq.~\eqref{eq:within-trial LLR update}, we define the prior after the $i$-th decision as:
        \begin{equation}
            y_0^{i+1}\left(y_n^i,d^i,r^i\right)=\ln\frac{\Pr\left(s^{i+1}=s_+\vert y_n^i,d^i,r^i\right)}{\Pr\left(s^{i+1}=s_-\vert y_n^i,d^i,r^i\right)}.
            \label{eq:across-trial LLR definition}
        \end{equation}
        Using the Law of Total Probability and conditional independence, we obtain the following update rule:
        \begin{equation}    y_0^{i+1}\left(y_n^i,d^i,r^i\right)=\text{sign}\left(d^i\right)\ln\begin{cases}
                \frac{\displaystyle (1-\epsilon)qe^{\vert y_n^i\vert}+\epsilon(1-q)}{\displaystyle \epsilon qe^{\vert y_n^i\vert}+(1-\epsilon)(1-q)}, & r^i=\circ\\
                \frac{\displaystyle (1-\epsilon)(1-q)e^{\vert y_n^i\vert}+\epsilon q}{\displaystyle \epsilon(1-q)e^{\vert y_n^i\vert}+(1-\epsilon)q}, & r^i=\times
            \end{cases},
            \label{eq:across-trial LLR update}
        \end{equation}
        where $\text{sign}\left(d^i\right)=\pm 1$ corresponds to $d^i=s_\pm$.
        Together, Eq.~\eqref{eq:within-trial LLR update} and Eq.~\eqref{eq:across-trial LLR update} determine a Bayesian agent's belief in our dynamic foraging task.   
        
    \subsection{Commitment Rules: Rewardmax and Infomax Models}
    
        Having established the Bayesian agent's belief-update rules, we now define the different commitment rules for the rewardmax (reward-maximizing) and infomax (information-maximizing) strategies under a fixed time step budget ($N$ time steps). 
        To construct these rules, we adopt a dynamic-programming perspective (Fig.~\ref{fig:task_schematics}\textbf{B}). 
        Specificallly, an agent with current belief $y_n^i$ chooses between two actions: 1) make a new observation, $\xi_{n+1}^i$, and update the belief to $y_{n+1}^i$, while the environment stays in the same state; or 2) commit to a decision, $d^i$, receive probabilistic feedback (reward), and update the belief to  $y_0^{i+1}$ based on the feedback, with the environmental state changing with probability $\epsilon$.
        Both of these actions (sampling and commitment, respectively) have an associated time-step cost, defined below.
        The action-selection process repeats until the agent exhausts their budget of $N$ time steps.
        To determine which action to take, the agent uses a belief-dependent utility function $U$, which depends on the agent's objective function, detailed below.
        
        \paragraph{Rewardmax Commitment Rule}
            To define the rewardmax utility function, we assume that the agent acts to optimize the average reward rate, $\rho$, over their whole time-step budget, which we define as 
            \begin{equation}
                \rho=\frac{\langle R\rangle}{\langle T_t\rangle +\langle T_i\rangle},
                \label{eq:reward_rate_definition}
            \end{equation}
            where $\langle R\rangle$ is the average reward, $\langle T_t\rangle$ is the average total time of a decision, and $\langle T_i\rangle$ is the average inter-decision interval~\cite{barendregt2022,drugowitsch2012}.
            All averages are taken across the entire $N$ time-step budget.
            The $N$ total time steps in the agent's budget restrict their ability to forage (e.g., limited seasonal availability of foraging resources).
            Our problem is thus a finite-horizon optimization, and as such the denominator of Eq.~\eqref{eq:reward_rate_definition} will always equal $N$. 
            Thus, only the terms in the numerator need to be maximized.
            
            To optimize the reward-based utility function, $U_{\text{rm}}$, we use Bellman's equation and backward induction on the time-step budget~\cite{barendregt2022,bellman1966,drugowitsch2015}.
            At the final time step, the agent has nothing to gain from making an observation. 
            Thus, the utility function at time step $N$ is determined by the utility of committing to a decision:
            \begin{equation}
                \begin{aligned}
                    U_{\text{rm}}\left(p_n^i;N\right)={}&\max\left\{U^+_{\text{rm}}\left(p_n^i;N\right),U^-_{\text{rm}}\left(p_n^i;N\right)\right\},\\
                    ={}&\max\begin{Bmatrix}
                        p_n^i\left[qR_++(1-q)R_-\right]+\left(1-p_n^i\right)\left[(1-q)R_++qR_-\right],\\
                        \left(1-p_n^i\right)\left[qR_++(1-q)R_-\right]+p_n^i\left[(1-q)R_++qR_-\right]
                    \end{Bmatrix},
                \end{aligned}
                \label{eq:rewardmax_utility_final_action}
            \end{equation}
            where $U^\pm_\text{rm}$ is the utility of committing to the decision $d^i=s_\pm$, $p_n^i=\Pr(s^i=s_+\vert\xi_{1:n}^i)=\frac{1}{1+e^{-y_n^i}}$ is the probability the $i$-th environmental state is $s_+$ given the sequence of observations $\xi_{1:n}^i$ (which we refer to as the state likelihood), and $R_\pm$ is the reward/punishment delivered if the feedback is $r^i=\pm$.
            Because the state likelihood, $p_n^i$, and the belief, $y_n^i$, are equivalent, we use the state likelihood in our utility function definitions for notational simplicity.
            
            With the final utility function constructed, we can now perform backward induction. 
            Consider time step number $k<N$, and assume that the agent's belief is $y_n^i$.
            In this case, we must modify the commitment utility functions $U^\pm_\text{rm}$ to reflect the utility of moving to the $(i+1)$-st environmental state. 
            If the agent chooses to commit, the new belief will be $y_0^{i+1}$.
            For $U^+_\text{rm}$, the new utility is given by:
            \begin{equation}
                \begin{aligned}
                    U_{\text{rm}}^+\left(p_n^i;k\right)={}&p_n^i\bigg[q\left\{R_++\gamma U_{\text{rm}}\left(p_0^{i+1}\left(p_n^i,s_+,+\right);k+\tau_d\right)\right\}\\
                    +&(1-q)\left\{R_-+\gamma U_{\text{rm}}\left(p_0^{i+1}\left(p_n^i,s_+,-\right);k+\tau_d\right)\right\}\bigg]\\
                    +&\left(1-p_n^i\right)\bigg[(1-q)\left\{R_++\gamma U_{\text{rm}}\left(p_0^{i+1}\left(p_n^i,s_+,+\right);k+\tau_d\right)\right\}\\
                    +&q\left\{R_-+\gamma U_{\text{rm}}\left(p_0^{i+1}\left(p_n^i,s_+,-\right)\right)\right\}\bigg],
                \end{aligned}
                \label{eq:rewardmax_U^+_general_action}
            \end{equation}
            where $\tau_d>0$ is the time step cost of commitment, effectively the delay period between potential foraging opportunities, and $\gamma\in[0,1]$ is a discounting term on future utility.
            As $\gamma\to0$, the model converges to a greedy (i.e., single-action optimal) strategy that does not take possible future utility gains into account. 
            We can obtain an expression for $U^-_\text{rm}$ following the same reasoning.
            
            We must also introduce the utility of sampling an additional piece of environmental evidence $U^s_\text{rm}$, which is equal to the average utility obtained via sampling~\cite{drugowitsch2012, barendregt2022}:
            \begin{equation}
                U_{\text{rm}}^s\left(p_n^i;k\right)=\gamma\left\langle U_{\text{rm}}\left(p_{n+1}^i;k+\tau_s\right)\vert p_n^i\right\rangle_{p_{n+1}^i}=\gamma\int_0^1U_{\text{rm}}\left(p_{n+1}^i;k+\tau_s\right)f_p\left(p_{n+1}^i\vert p_n^i\right)\,dp_{n+1}^i,
                \label{eq:rewardmax_U^s_general_action}
            \end{equation}
            where $\tau_s>0$ is the time step cost of sampling, or the delay period between evidence-gathering opportunities, and $f_p$ is the conditional distribution of obtaining the future state likelihood $p_{n+1}^i$ given the current likelihood $p_n^i$.
            This conditional distribution depends on the evidence-generating distribution $f_\pm(\xi)$ (for a Gaussian-distributed example, see~\cite{barendregt2022}).
            For our task, we assume that the observations of the environmental state, $\xi^i_j,$ are Bernoulli-distributed with parameter $h$ so that:
            \begin{equation}
                f_\pm(\xi)=\begin{cases}
                    h, & \xi=\pm 1\, \land\, s^i=s_\pm\\
                    1-h, & \xi=\mp 1\, \land\, s^i=s_\pm
                \end{cases}.
                \label{eq:Bernoulli_evidence_distribution}
            \end{equation}
            Given this evidence distribution, the conditional state likelihood distribution $f_p$ is
            \begin{equation}
                f_p\left(p_{n+1}^i\vert p_n^i\right)=\begin{cases}
                    p_n^ih+\left(1-p_n^i\right)(1-h), & p_{n+1}^i=\frac{\displaystyle hp_n^i}{\displaystyle hp_n^i+(1-h)\left(1-p_n^i\right)}\\
                    p_n^i(1-h)+\left(1-p_n^i\right)h, & p_{n+1}^i=\frac{\displaystyle (1-h)p_n^i}{\displaystyle (1-h)p_n^i+h\left(1-p_n^i\right)}
                \end{cases},
                \label{eq:Bernoulli_evidence_belief_transfer}
            \end{equation}
            which can be obtained via a stochastic change of variables. 
            This distribution describes the conditional probability that a subsequent sample will update the belief to a particular value, given the current belief.
            
            Given Eq.~\eqref{eq:rewardmax_U^+_general_action} and Eq.~\eqref{eq:rewardmax_U^s_general_action}, we can define the general rewardmax utility function as:
            \begin{equation}
                U_{\text{rm}}\left(p_n^i;k\right)=\max\left\{U_{\text{rm}}^+\left(p_n^i;k\right),U_{\text{rm}}^-\left(p_n^i;k\right),U_{\text{rm}}^s\left(p_n^i;k\right)\right\}.
                \label{eq:rewardmax_utility_general_action}
            \end{equation}
            At every time step, an ideal (Bayesian) foraging agent calculates this utility function given their current belief (or, equivalently, the state likelihood) and takes the action that has maximal utility.
            
        \paragraph{Infomax Commitment Rule}
            As in the traditional infotaxis model~\cite{vergassola2007}, the infomax model assumes that the agent seeks to minimize the uncertainty about the environmental state as described by the state entropy~\cite{cover1999}: 
            \begin{equation}
                \begin{aligned}
                    H\left(p_n^i\right)={}&-\sum_{s\in s_\pm}\Pr\left(s^i=s\vert\xi_{1:n}\right)\log_2\Pr\left(s^i=s\vert \xi_{1:n}\right).\\
                    ={}&-p_n^i\log_2\left(p_n^i\right)-\left(1-p_n^i\right)\log_2\left(1-p_n^i\right).
                \end{aligned}
                \label{eq:state_entropy}
            \end{equation}
            Using Eq.~\eqref{eq:state_entropy}, we can again apply Bellman's equation and backward induction, using uncertainty minimization as the objective function, to derive the infomax utility function.
            Unlike standard infotaxis, which is formulated for static or slowly varying environments to maximize expected information gain at each time step, our infomax utility function optimizes information gain over a future sequence of environmental states under the agent's budget of $N$ time steps.
            This approach allows us to establish a proper comparison with the rewardmax model.
            By introducing discounting on future utility and setting $\gamma = 0$, we can recover the traditional infotaxis objective that accounts for only the next time step.
            
            As with the rewardmax utility function, we start by considering the final time step in the agent's budget.
            At this point, future information gains are not possible (i.e., $U_{\text{im}}\left(p_n^i;N\right)=0$).
            For time step number $k<N$, we can calculate the utility $U_{\text{im}}^+(p_n^i;k)$ of a decision $d^i=s_+$, when the state likelihood is $p_n^i$ based on the information obtained from the probabilistic feedback $r^i$.
            However, this information gain is not simply the state entropy of $p_0^{i+1}$, because even if feedback is completely unreliable and decoupled from the agent's performance (i.e.,~$q=0.5$), the previous state ($s^i$) still partially predicts the next ($s^{i+1}$) as long as $\epsilon \neq 0.5$.
            Therefore, we must normalize the information gain using the case of non-informative feedback, $q=0.5$, as a baseline:
            \begin{equation}
                \begin{aligned}
                    U_{\text{im}}^+\left(p_n^i;k\right)={}&H\left(p_n^i+\epsilon-2p_n^i\epsilon\right)\\
                    -{}&\left[qp_n^i+(1-q)\left(1-p_n^i\right)\right]H\left(p_0^{i+1}\left(p_n^i,s_+,+\right)\right)\\
                    -{}&\left[q\left(1-p_n^i\right)+(1-q)p_n^i\right]H\left(p_0^{i+1}\left(p_n^i,s_+,-\right)\right)\\
                    +{}&p_n^i\left[q\gamma U_{\text{im}}\left(p_0^{i+1}\left(p_n^i,s_+,+\right);k+\tau_d\right)+(1-q)\gamma U_{\text{im}}\left(p_0^{i+1}\left(p_n^i,s_+,R_i\right);k+\tau_d\right)\right]\\
                    +{}&\left(1-p_n^i\right)\left[(1-q)\gamma U_{\text{im}}\left(p_0^{i+1}\left(p_n^i,s_+,+\right);k+\tau_d\right)+q\gamma U_{\text{im}}\left(p_0^{i+1}\left(p_n^i,s_+,-\right);k+\tau_d\right)\right],
                \end{aligned}
                \label{eq:infomax_U^+_general_action}
            \end{equation}
            where the expression $p_n^i+\epsilon+2p_n^i\epsilon$ is the simplified form of $p_0^{i+1}$ from Eq.~\eqref{eq:across-trial LLR update} when $q=0.5$.
            As with the reward-maximizing strategy, the utility $U_{\text{im}}^-$ for a decision $d^i=s_-$ can be obtained in the same manner.
            
            We then calculate the utility of making another observation as the reduction in uncertainty expected from updating the belief about the current state, $s^i$, based on that observation:
            \begin{equation}
                \begin{aligned}
                    U_{\text{im}}^s\left(p_n^i;k\right)={}&H\left(p_n^i\right)-\left\langle H\left(p_{n+1}^i\right)\vert p_n^i\right\rangle_{p_{n+1}^i}+\gamma\left\langle U\left(p_{n+1}^i;k+\tau_s\right)\vert p_n^i\right\rangle_{p_{n+1}^i}.\\
                    ={}&H\left(p_n^i\right)-\int_0^1\left[H\left(p_{n+1}^i\right)-\gamma U\left(p_{n+1}^i;k+\tau_d\right)\right]f_p\left(p_{n+1}^i\vert p_n^i\right)\,dp_n^i,
                \end{aligned}
                \label{eq:infomax_U^s_general_action}
            \end{equation}
            where the conditional distribution $f_p$ is the same as the distribution given by Eq.~\eqref{eq:Bernoulli_evidence_belief_transfer}.
            Combining Eq.~\eqref{eq:infomax_U^+_general_action} and Eq.~\eqref{eq:infomax_U^s_general_action} gives the general infomax utility function:
            \begin{equation}
                U_{\text{im}}\left(p_n^i;k\right)=\max\left\{U_{\text{im}}^+\left(p_n^i;k\right),U_{\text{im}}^-\left(p_n^i;k\right),U_{\text{im}}^s\left(p_n^i;k\right)\right\}.
            \end{equation}
            
\section{Results}
    We compared how the rewardmax and infomax strategies perform our dynamic foraging task.
    The two strategies exhibit similar, but distinguishable, behavioral patterns under different task conditions governed by changes in environmental stability, $\epsilon$, and feedback reliability, $q$. 
    These differences reflect the overall benefits of rewardmax for obtaining more reward, on average, but of infomax for obtaining rewards more reliably, and go beyond standard anlayses that focus on mean rewards.  
    
    \subsection{Rewardmax and Infomax Strategies Prescribe Similar Explore-Exploit Trade-Offs}        
\begin{figure}[t!]
            \centering
            \includegraphics[width=\linewidth]{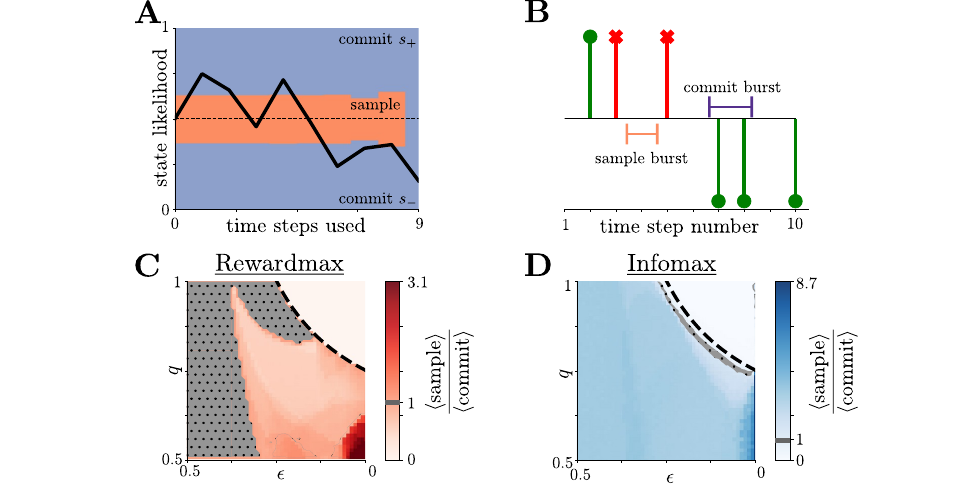}
            \caption{\textbf{Both Strategies Predict Explore-Exploit Phase Transitions.}
                \textbf{A,B:} Example realization of the reward-maximizing strategy in state-likelihood space $p_n^i$ (\textbf{A}) and action space (\textbf{B}).
                In \textbf{B}, upward (downward) stems denote decisions towards $s_+$ ($s_-$), and green stems with circles (red spines with strikes) denote decisions that are rewarded (punished).
                Action numbers without stems denote actions for which the agent samples the environment for evidence.
                From each realization, we extract two behavioral metrics: 1) the average number of sequential commitments within a commit burst (e.g., purple), and 2) the average number of sequential samples within a sample burst (e.g., orange).
                In this particular realization, there are four commit bursts of lengths 2, 1, 2, and 1, and there are four sample bursts of length 1.
                \textbf{C:} Ratio of average sample burst length to average commit burst length for rewardmax behavior as functions of environmental stability ($\epsilon$) and reward reliability ($q$).
                In the realization shown in \textbf{B}, the average sample burst length is 1, and the average commit length is 1.5, resulting in a ratio metric of $\frac{2}{3}$.
                In the gray, dotted region of parameter space this ratio is unity. 
                Values larger than unity indicate more exploratory behavior, whereas values smaller than unity indicate more exploitative behavior.
                Black, dashed line shows approximate location of the phase-transition boundary given by Eq.~\eqref{eq:phase_transition_approximation_parameter_regions}.
                As the environment becomes more stable ($\epsilon$ decreases) and provides more reliable feedback ($q$ increases), the agent opts to never sample, relying on pure exploitation that reliably delivers reward and evidence.
                Results generated with $10^4$ realizations, each with Bernoulli-evidence parameter $h=0.75$, time step budget $N=10$, action time step costs $(\tau_d,\tau_s)=(1,1)$, reward structure $\left(R_c,R_i\right)=(100,\text{-}100)$, and no discounting ($\gamma=1$).
                \textbf{D:} Same as \textbf{C}, but for infomax behavior.
                Infomax behavior exhibits a similar, but less sharp, phase transition to pure exploitation than rewardmax.}
            \label{fig:explore_exploit_phase_transition}
        \end{figure}

        Single stochastic simulations of the rewardmax strategy on our dynamic foraging task (Fig.~\ref{fig:explore_exploit_phase_transition}\textbf{A}) show that the regions of belief space where either commitment or sampling is the optimal action vary non-trivially as the agent uses actions from their budget.
        From these individual realizations, we can extract two behavioral metrics (schematized in Fig.~\ref{fig:explore_exploit_phase_transition}\textbf{B}): 1) the number of commitment actions within a sequence of consecutive commitments, which we call a ``commit burst,'' and 2) the number of sample actions within a sample sequence, which we call a ''sample burst.''
        Both of these burst lengths are non-negative integers, because, for instance, even a single commitment action that is both preceded and followed by sampling actions counts as a sample burst of length one (see  Fig.~\ref{fig:explore_exploit_phase_transition}).
        The statistics of these two quantities indicate the agent's preference for resource exploitation and environmental exploration, respectively.
        Previous work in foraging and psychology \cite{addicott2017} has demonstrated that agents often navigate a trade-off between these two stereotypical behaviors, as in the classic ``win-stay, lose-switch'' strategy \cite{ma2022, robbins1952}.
		
        By manipulating the environmental change probability, $\epsilon$, and feedback reliability, $q$, we find that both rewardmax and infomax behaviors undergo a phase transition (Fig.~\ref{fig:explore_exploit_phase_transition}\textbf{C,D}). 
        Specifically, as $\epsilon$ decreases and $q$ increases, both strategies prescribe a transition from sampling evidence at least occasionally (exploring) to never sampling evidence and committing to a decision with every action (exploitation; Fig.~\ref{fig:explore_exploit_phase_transition}\textbf{B,C}).
        As these environmental parameters approach this phase transition, the reliability of feedback and the stability of the environment both increase, allowing the agent to carry a stronger belief into the next trial.
        In these environments, agents maximizing reward value commitment because it offers both immediate reward and information about future rewards.
        Agents maximizing information value commitment because the resulting feedback obtained from the probabilistic reward carries more information, on average, than observations of the environment.
        Accordingly, the phase transition's boundary can be approximated by comparing the average change in belief magnitude from feedback versus sampling, given that the agent believes either state is equally likely (i.e.,~for LLR belief $y=0$) as the level set
        \begin{equation}
            (1-h)\left[(1-\epsilon)q+\epsilon(1-q)\right]=h\left[\epsilon q+(1-\epsilon)(1-q)\right].
            \label{eq:phase_transition_approximation_parameter_regions}
        \end{equation}
        For details of how to obtain this approximation, see Appendix~\ref{app:phase_transition_approximation}.
        Once feedback increments belief more than sampling, both strategies generally prefer to always commit.
        
        The phase transition's boundary shifts in predictable ways as other task parameters are changed (see Fig.~\ref{supfig:rewardmax_phase_transition}-\ref{supfig:infomax_phase_transition} in Appendix~\ref{app:supplemental_figures}). 
        For example, if the commit time cost $\tau_d$ is larger than the sample time cost $\tau_s$, the exploitation region shrinks, because commitment must produce a larger benefit to balance out the larger cost. 
        The opposite holds true when $\tau_s>\tau_d$; i.e., the exploitation region grows.
        Similarly, the exploitation region shrinks as the environmental evidence's Bernoulli parameter $h$ increases, because environmental evidence carries more information than information from reward.
        The opposite holds true when $h$ decreases; i.e., the exploitation region grows.
        Furthermore, if the magnitude of reward $\vert R_c\vert$ is larger than the magnitude of punishment $\vert R_i\vert$, the rewardmax model's exploitation region expands, because commitment produces a higher average reward utility. 
        The opposite holds true  when $\vert R_i\vert>\vert R_c\vert$ (i.e., the exploitation region shrinks). 
        The infomax model's exploitation region does not change with reward or punishment, because information utility operates independently from reward. 
        The phase transition for both models does not appear to have a noticeable dependence on the agent's time-step budget $N$.
		
        \begin{figure}[t]
            \centering
            \includegraphics[width=\linewidth]{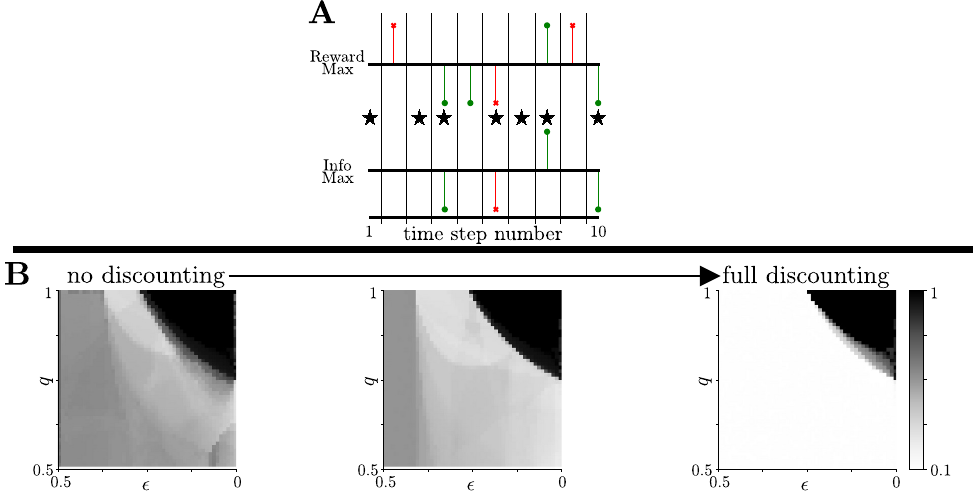}
            \caption{\textbf{Action Alignment of Rewardmax and Infomax Behavior.}
                \textbf{A:} Schematic of the alignment metric. 
                For both models, we generate an action-space representation (using the same notation as in Fig.~\ref{fig:explore_exploit_phase_transition}\textbf{B}) from a common environmental belief realization. 
                We quantified the similarity of behaviors generated by the two models as ``action alignment,'' defined as the proportion of identical actions (marked with stars) when presented with the same belief about the current environmental state.
                \textbf{B:} Alignment of rewardmax and infomax behaviors as a function of environmental stability ($\epsilon$) and reward reliability ($q$) and progressing from no temporal discounting (left) to full temporal discounting of reward (right).
                Strategies are most distinct (i.e., alignment is lowest) in environments with intermediate stability (moderate $\epsilon$) and high reliability (large $q$).
                This distinctness is emphasized as future utility is increasingly discounted ($\gamma\to0$).
                Results generated with $10^4$ realizations of common belief trajectories with $\gamma\in\{1,0.5,0\}$ (left-to-right panels) and all other task parameters as in Fig.~\ref{fig:explore_exploit_phase_transition}.}
            \label{fig:action_alignment}
        \end{figure}
        Despite these similar, task-dependent phase transitions between exploitative and exploratory behaviors, there are noticeable behavioral differences in the way the two strategies perform the tasks. 
        Using the same experimental parameter space, we define ``action alignment''  between the strategies as the fraction of times that both prescribe the same action, given the same environmental belief (see Fig.~\ref{fig:action_alignment}\textbf{A} for a schematic). 
        Action alignment is highest when $\epsilon$ is small and $q$ is large, corresponding to environments in which commitment reliably gives a high level of information (Fig.~\ref{fig:action_alignment}\textbf{B}), so that both strategies choose commitment is chosen on every time step.
        However, the strategies differ across the remainder of parameter space, with lowest alignment when $\epsilon$ is at an intermediate level and $q$ is large. 
        This difference increases as future utility is more strongly discounted ($\gamma\to0$), with the two strategies producing different behaviors across all but the most exploitation-dominant task conditions when only immediate rewards are considered.

        \begin{figure}[t]
            \centering
            \includegraphics[width=\linewidth]{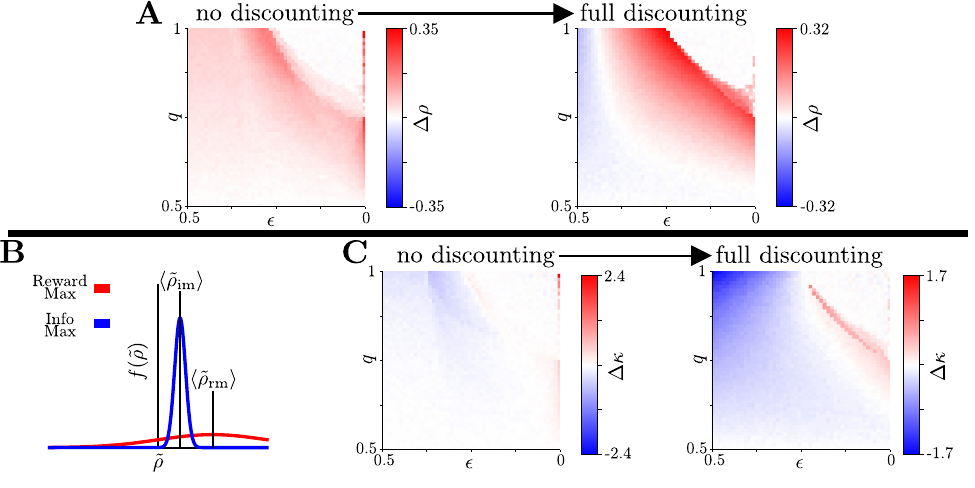}
            \caption{\textbf{Robustness of the Infomax Model.}
                \textbf{A:} Average normalized reward rate differential $\Delta\rho=\frac{\rho_{\text{rm}}-\rho_{\text{im}}}{R_c}$ as a function of environmental stability, $\epsilon$, and reward reliability, $q$,  progressing from no temporal discounting ($\gamma=1$, left) to full temporal discounting ($\gamma=0$, right) of reward.
                \textbf{B:}~Schematic of reward rate distributions for both models.
                The rewardmax strategy tends to maximize the reward obtained, on average, but often with more variability than the infomax strategy across repeated instances of the same task conditions. 
                Infomax thus guards against precipitously low reward returns by producing consistently adequate rewards.
                \textbf{C:} Robustness differential $\Delta\kappa=\kappa_{\text{rm}}-\kappa_{\text{im}}$, with a model's robustness defined as $\kappa=\frac{\langle\tilde{\rho}\rangle}{\text{std}(\tilde{\rho})}$.
                Even as rewardmax yields larger average reward rates, infomax delivers a more narrow distribution of reward rates, making it a more robust strategy.
                This robustness advantage is enhanced as discounting on future utility increases.}
            \label{fig:performance_robustness}
        \end{figure}
        
    \subsection{Infomax Produces More Robust Rewards Than Rewardmax}
        We further explored how the two strategy classes performed in terms of obtaining reward as their levels of temporal discounting were varied.
        Over an ensemble of belief realizations, both the rewardmax and infomax models produce distributions, $f_{\text{rm}}\left(\tilde{\rho}\right)$ and $f_{\text{im}}\left(\tilde{\rho}\right)$, respectively, over empirical reward rates $\tilde{\rho}=\frac{R}{T_t+T_i}$.
        We examined not simply the average, $\langle\tilde{\rho}\rangle$ (or, equivalently, $\rho$ as defined in Eq.~\eqref{eq:reward_rate_definition}), but the entire distribution of reward outcomes for each environment and strategy.
        
        As expected, the rewardmax model tends to outperform the infomax model in terms of average reward obtained across task conditions (by $\sim25-33\%$ of a correct response; Fig.~\ref{fig:performance_robustness}\textbf{A}). However, this performance advantage is more nuanced than one might expect.
        First, the advantages are most prominent in environments with a combination of intermediate-to-high reward reliability ($q$) and stability ($\epsilon$). 
        In contrast, both strategies lead to nearly identical average reward rates when reliability is low and when purely exploitative behaviors predominate for both models. 
        Second, as future utility is discounted ($\gamma\to0$), the infomax model increasingly produces larger average reward rates than the rewardmax model in highly volatile and/or unreliable environments.
        This effect can be explained by the fact that, as $\gamma\to0$, the rewardmax model transitions to a greedy strategy that makes only commitments, ignoring potential future gains of sampling, whereas the infomax model continues to balance sampling-based and response feedback-based evidence gathering, making it a more robust strategy across objectives.
        This difference, coupled with a reward structure that includes punishment for incorrect decisions, allows the infomax model to narrow the performance advantage.
        
        In addition to these differences in average reward rates, the two strategies produce striking differences in the reliability with which those rewards are obtained (Fig.~\ref{fig:performance_robustness}\textbf{B}). 
        Broad distributions offer substantial probability of low performance, even if they have higher average reward rates. 
        In contrast, narrow distributions guard against the probability of low performance, even if their average reward rates are slightly lower. 
        We defined a robustness metric, $\kappa$, as: 
        \begin{equation}
            \kappa=\frac{\langle\tilde{\rho}\rangle}{\text{std}(\tilde{\rho})},
            \label{eq:robustness_metric}
        \end{equation}
        which favors models that deliver a large average reward rate (which can be close to zero in our task and thus is in the numerator) as well as a consistent reward rate (the denominator). 
        Robustness is consistently higher for the infomax than the rewardmax strategy, across a broad range of task conditions and particularly as future utility discounting increases (Fig.~\ref{fig:performance_robustness}\textbf{C}).
        These results demonstrate reasons for an infomax strategy to be preferred over a rewardmax strategy in naturalistic environments that carry risk.
        
\section{Discussion}
    In this work, we provided a detailed analysis of how reward-maximizing (rewardmax) and information-maximizing (infomax) strategies perform a sequential, dynamic extension of a classic foraging task. 
    We showed that both strategies exhibit a similar explore-exploit trade-off, with more exploring in volatile environments and more exploiting in stable environments.
    We also showed that these strategies can be distinguished from each other by their action sequences in non-extremal environments.
    These results complement previous comparisons of normative and heuristic strategies, which found that the largest performance differences between different decision strategies occur in environments with intermediate levels of uncertainty~\cite{barendregt2019,tavoni2022}.
    Finally, we showed that a rewardmax model often provides a performance advantage in average reward rate, but an infomax model dominates in producing a consistent distribution of nearly optimal reward rates.
    
    These results demonstrate the practical utility of information seeking in naturalistic foraging settings with risk and provide several testable hypotheses for further study. 
    First, we predict that naturalistic behaviors should exhibit a sharp transition from exploration to exploitation at a threshold between modest and high environmental stability and feedback reliability. 
    Identifying these abrupt shifts in strategy could clarify how organisms adapt to changing environments. 
    Second, as reward reliability decreases or environmental volatility increases, reliable long-term reward outcomes depend more on favoring information gathering, although potentially at the cost of short-term gains.
    This contrast underscores the trade-off between short-term exploitation and long-term adaptation, highlighting the conditions under which information-seeking behaviors provide the greatest advantage.
	
    Although our work focused on decision making in the context of a foraging task, our findings likely generalize to other problems. 
    In particular, our task shares features with those modeled by partially observed Markov decision processes (POMDPs) in theories of economics, planning, and machine learning~\cite{auer2002, bergemann2006, kocsis2006}. 
    Stationary POMDPs (i.e., those designed to operate in statistically stable environments) have been compared to traditional infotaxis and reward-maximizing strategies~\cite{loisy2022,heinonen2023optimal}. 
    Our work shows that those comparisons can miss many of the rich differences in behavior that those strategies can produce in more dynamic environments. 
    Our work also implies that future studies of how these different strategies compare to each other should carefully consider the effects of future reward discounting, which is often implemented poorly in planning-type experiments because of computational costs~\cite{callaway2022,ho2022}.
    Based on our results, future utility discounting can dramatically alter the relative merits of reward- versus information-seeking strategies.
	
    We aimed to construct a simple sequential, dynamic foraging environment and feedback structure, so we could link observed behavior directly to the agent's model selection.
    However, there are several natural extensions of our task that could yield further insights.
    The first would be to consider environments with state changes that can occur within a single deliberative process, so the belief must adapt to state fluctuations between commitments.
    This type of dynamic environment has been studied for non-sequential decision tasks~\cite{barendregt2019,glaze2015,veliz2016}. 
    We hypothesize that such environments would likely engender additional differences in rewardmax versus  infomax behaviors, given the even more extensive need to learn about the statistics of the within-trial change process to inform the decisions.
    
    A second possible extension would be to consider multiple rewarding alternatives beyond the two we considered.
    For our two-alternative task, the information gained by mutual exclusivity allowed both models to perform well across environmental parametrizations. 
    This mutual exclusivity is broken when there are more than two possible environmental states and evidence against one alternative increases the probability of multiple other alternatives.
    We hypothesize that under these conditions an infomax strategy has even more extensive benefits over a rewardmax strategy, given our finding that infomax tends to perform well in difficult, risky environments.
    We also predict that an infomax strategy may exhibit complex commitment patterns, because committing to, and obtaining feedback from, different alternatives in sequence can yield significant information, even if feedback indicates an incorrect decision.
    These nuanced commitment behaviors may prove useful in studying various choice paradoxes observed in cognitive psychology, such as the independence of irrelevant alternatives~\cite{luce1959,shafir2002,tajima2019}.

    Overall, our results demonstrate that while reward-seeking strategies maximize average gains, information-seeking strategies provide greater consistency and robustness in uncertain environments. 
    Infomax’s ability to generate more stable reward distributions suggests that organisms operating in dynamic settings may benefit from prioritizing information acquisition, even at the cost of occasional missed immediate rewards. 
    These findings highlight the adaptive value of information-seeking behaviors for future gains, particularly in volatile environments in which reward contingencies change unpredictably. 
    In other words, the pursuit of information may be as crucial to survival as the pursuit of resources themselves.
    
\subsection*{Acknowledgments} 
    NWB, JIG, KJ, and ZPK were all supported by a CRCNS grant (NSF DMS-2207700).
    ZPK was supported by NIH BRAIN 1R01EB029847-01. 
    KJ was supported by NSF-DBI-1707400 and NIH RF1MH130416.

\subsection*{Code Availability} 
    See \url{https://github.com/nwbarendregt/SequentialRewardInfo} for the Python code used to generate all results and figures.
    \bibliography{references}
\appendix
\section{Phase Transition Approximation}
    \label{app:phase_transition_approximation}
    Here we detail the calculation for approximating the explore-exploit phase transition's location as shown in Fig.~\ref{fig:explore_exploit_phase_transition}.
    For our approximation, we compare the average change in belief magnitude from feedback versus sampling and assume that the agent takes whichever action leads to the largest magnitude change when either state is equally likely (i.e., for LLR belief $y=0$).
    This approximation can be thought of as choosing the most ``impactful'' action on the very first action of the sequence, and if it is most impactful for the agent to commit to a decision, they will likely continue making commitments for the remainder of their action sequence. 
    Throughout these calculations, we use the notation $Y_n^i$ to represent the random variable for the agent's LLR belief after observing $n$ environmental samples in the $i$-th environment, and $y_n^i$ to represent the realized LLR value, similar to the notation used throughout Section~\ref{sec:methods}.
    
    To calculate the average LLR magnitude change associated with feedback, we start by using the Law of Total Probability to condition the magnitude change over the decision $d^i$ and feedback $r^i$:
    \begin{equation}
        \mathbb{E}\left[\vert Y_0^{i+1}-Y_n^i\vert\vert Y_n^i=y_n^i=0\right]=\sum_{d^i\in\left\{s_+,s_-\right\},r^i\in\left\{\circ,\times\right\}}\mathbb{E}\left[\vert Y_0^{i+1}\vert\vert y_n^i=0,d^i,r^i\right]\Pr\left(d^i,r^i\vert y_n^i=0\right).
        \label{eq:LLR_magnitude_feedback_definition}
    \end{equation}
    Next, we note that, given the conditioning on $y_n^i$, $d^i$, and $r^i$, the LLR belief $Y_0^{i+1}$ is fixed and given by Eq.~\eqref{eq:across-trial LLR update}.
    Inspecting Eq.~\eqref{eq:across-trial LLR update}, we can see that the LLR belief $Y_0^{i+1}$ is the same regardless of $r^i$ and $d^i$, allowing us to pull this expectation out of the sum in Eq.~\eqref{eq:LLR_magnitude_feedback_definition}. 
    At this point, the remaining sum of probabilities equals unity, because the sum contains all possible realizations of the joint event $(r^i,d^i)$.
    Performing these simplifications yields the expected change in LLR magnitude:
    \begin{equation}
        \mathbb{E}\left[\vert Y_0^{i+1}-Y_n^i\vert\vert Y_n^i=y_n^i=0\right]=\Bigg\vert\ln\frac{(1-\epsilon)q+\epsilon(1-q)}{\epsilon q+(1-\epsilon)(1-q)}\Bigg\vert.
        \label{eq:eq:LLR_magnitude_feedback_simplified}
    \end{equation}

    To calculate the expected change in LLR magnitude associated with environmental sampling, we first define this magnitude change using Eq.~\eqref{eq:within-trial LLR update} as:
    \begin{equation}
        \vert Y_{n+1}^i-Y_n^i\vert=\Bigg\vert\ln\frac{f_+(\xi_{n+1})}{f_-(\xi_{n+1})}\Bigg\vert.
        \label{eq:LLR_magnitude_sampling_definition}
    \end{equation}
    Using the fact that $\xi_{n+1}$ follows the Bernoulli distribution given by Eq.~\eqref{eq:Bernoulli_evidence_distribution} allows us to simplify this magnitude change:
    \begin{equation*}
        \vert Y_{n+1}^i-Y_n^i\vert=\begin{cases}
            \big\vert \ln\frac{\displaystyle h}{\displaystyle 1-h}\big\vert, & \xi_{n+1}=+1\\
            \big\vert\ln\frac{\displaystyle 1-h}{\displaystyle h}\big\vert, & \xi_{n+1}=-1
        \end{cases},
    \end{equation*}
    which, due to logarithm rules, is constant for all possible realizations of $\xi_{n+1}$ and is independent of the agent's current LLR belief value:
    \begin{equation}
        \vert Y_{n+1}^i-Y_n^i\vert= \bigg\vert\ln\frac{h}{1-h}\bigg\vert.
        \label{eq:LLR_magnitude_sampling_simplified}
    \end{equation}

    From Eq.~\eqref{eq:eq:LLR_magnitude_feedback_simplified} and Eq.~\eqref{eq:LLR_magnitude_sampling_simplified}, we obtain our approximation: the phase transition occurs when the expected LLR magnitude change from feedback exceeds the expected LLR magnitude change from sampling, which implies the phase transition's boundary is where these two quantities are equal:
    \begin{equation}
        \Bigg\vert\ln\frac{(1-\epsilon)q+\epsilon(1-q)}{\epsilon q+(1-\epsilon)(1-q)}\Bigg\vert=\Bigg\vert \ln\frac{h}{1-h}\Bigg\vert.
        \label{eq:phase_transition_approximation_general_parameters}
    \end{equation}
    For the limits of environmental parameters ($\epsilon\in[0,0.5]$, $q\in[0.5,1]$) and evidence-generating parameters ($h\in[0.5,1]$) we considered, both logarithms in Eq.~\eqref{eq:phase_transition_approximation_general_parameters} output non-negative values, allowing us to make a further simplification by dispensing of the absolute values, producing Eq.~\eqref{eq:phase_transition_approximation_parameter_regions}:
    \begin{equation*}
        (1-h)\left[(1-\epsilon)q+\epsilon(1-q)\right]=h\left[\epsilon q+(1-\epsilon)(1-q)\right].
    \end{equation*}
    \pagebreak
\section{Supplemental Figures}
\label{app:supplemental_figures}
\renewcommand{\thefigure}{Sup-\arabic{figure}}
\setcounter{figure}{0}
\begin{figure}[ht!]
    \centering
    \includegraphics[width=\linewidth]{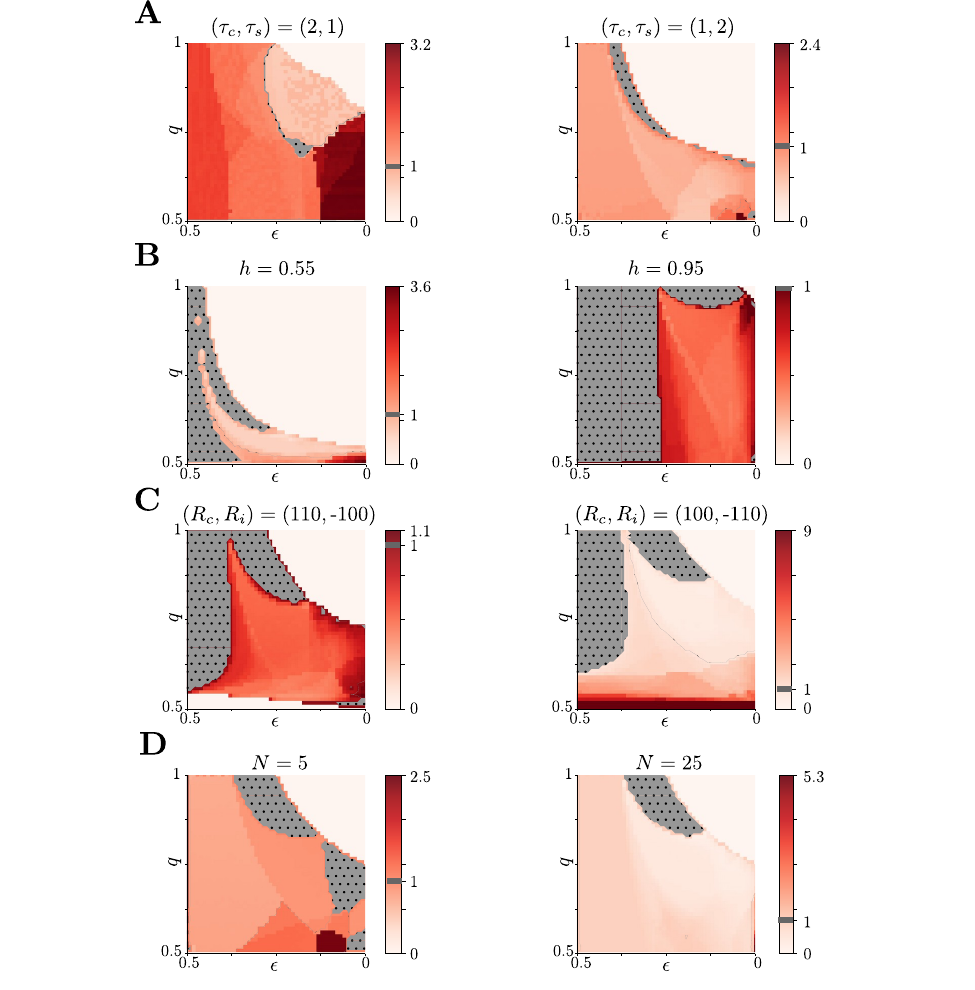}
    \caption{\textbf{Rewardmax Phase Transition Behavior.}
        \textbf{A:} Ratio of average sample burst length to average commit burst length for rewardmax behavior as in Fig.~\ref{fig:explore_exploit_phase_transition}\textbf{C}, but with action time step costs $(\tau_c,\tau_s)=(2,1)$ (left) and $(\tau_c,\tau_s)=(1,2)$ (right).
        \textbf{B:} Same as \textbf{A}, but for environmental evidence Bernoulli parameter $h=0.55$ (left) and $h=0.95$ (right).
        \textbf{C:} Same as \textbf{A}, but for reward structure $(R_c,R_i)=(110,-100)$ (left) and $(R_c,R_i)=(100,-110)$ (right).
        \textbf{D:} Same as \textbf{A}, but for action budget $N=5$ (left) and $N=25$ (right).
        Unless specifically altered, all other task parameters are the same as in Fig.~\ref{fig:explore_exploit_phase_transition}.}
    \label{supfig:rewardmax_phase_transition}
\end{figure}

\begin{figure}[ht!]
    \centering
    \includegraphics[width=\linewidth]{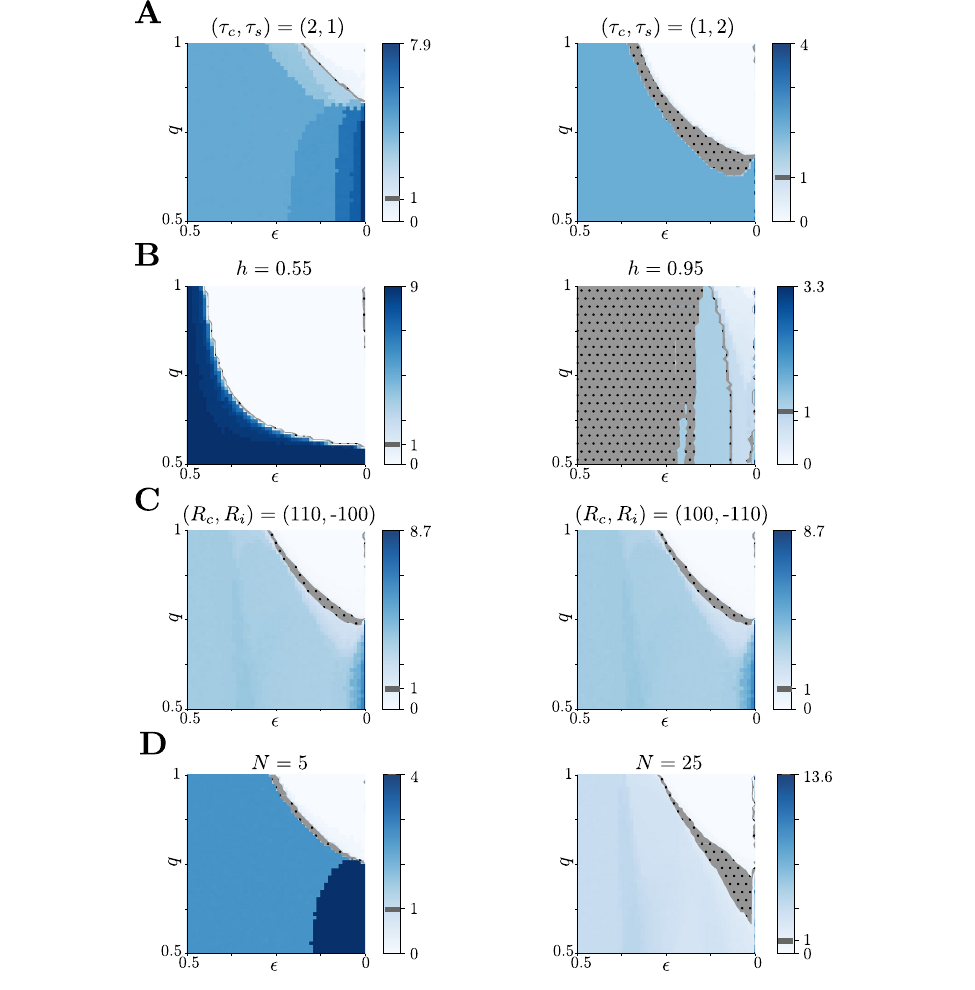}
    \caption{\textbf{Infomax Phase Transition Behavior.}
        \textbf{A:} Ratio of average sample burst length to average commit burst length for infomax behavior as in Fig.~\ref{fig:explore_exploit_phase_transition}\textbf{D}, but with action time step costs $(\tau_c,\tau_s)=(2,1)$ (left) and $(\tau_c,\tau_s)=(1,2)$ (right).
        \textbf{B:} Same as \textbf{A}, but for environmental evidence Bernoulli parameter $h=0.55$ (left) and $h=0.95$ (right).
        \textbf{C:} Same as \textbf{A}, but for reward structure $(R_c,R_i)=(110,-100)$ (left) and $(R_c,R_i)=(100,-110)$ (right).
        \textbf{D:} Same as \textbf{A}, but for action budget $N=5$ (left) and $N=25$ (right).
        Unless specifically altered, all other task parameters are the same as in Fig.~\ref{fig:explore_exploit_phase_transition}.}
    \label{supfig:infomax_phase_transition}
\end{figure}
\end{document}